\documentclass[letterpaper, 10 pt, conference]{ieeeconf}  
\IEEEoverridecommandlockouts
\usepackage{url}
\usepackage{times}
\usepackage{epsfig}
\usepackage{graphicx}
\usepackage{amsmath}
\usepackage{amssymb}
\usepackage{multirow}
\usepackage{makecell}
\usepackage{hyperref}

\usepackage{xcolor}
\usepackage{float}
\usepackage{lipsum}

\newcommand{\tabincell}[2]{\begin{tabular}{@{}#1@{}}#2\end{tabular}}

\title{\LARGE \bf
Transferable Active Grasping and Real Embodied Dataset}

\author{Xiangyu Chen$^{*}$, Zelin Ye$^{*}$, Jiankai Sun, Yuda Fan, Fang Hu, Chenxi Wang, and Cewu Lu$^{\dagger}$
\thanks{$^{*}$ These authors contribute equally to the work.}
\thanks{$^{\dagger}$ Cewu Lu is the corresponding author.}
\thanks{All authors are with the Department of Computer Science at Shanghai Jiao Tong University, Shanghai, China. \texttt{cxy\_1997 $\mid$ h\_e\_r\_o $\mid$ jiankai $\mid$ kurodakanbei $\mid$ hu-fang $\mid$ wcx1997 $\mid$ lucewu @ sjtu.edu.cn}}
}

\begin{document}
\maketitle

\begin{abstract}
Grasping in cluttered scenes is challenging for robot vision systems, as detection accuracy can be hindered by partial occlusion of objects. 
We adopt a reinforcement learning (RL) framework and 3D vision architectures to search for feasible viewpoints for grasping by the use of hand-mounted RGB-D cameras. 
To overcome the disadvantages of photo-realistic environment simulation, we propose a large-scale dataset called \emph{Real Embodied Dataset} (RED), which includes full-viewpoint real samples on the upper hemisphere with \emph{amodal annotation} and enables a simulator that has \textbf{real visual} feedback. Based on this dataset, a practical 3-stage transferable active grasping pipeline is developed, that is adaptive to unseen clutter scenes. In our pipeline, we propose a novel mask-guided reward to overcome the sparse reward issue in grasping and ensure category-irrelevant behavior. 
The grasping pipeline and its possible variants are evaluated with extensive experiments both in simulation and on a real-world UR-5 robotic arm.
\end{abstract}

\section{INTRODUCTION}

Reliable robotic grasping of a certain object from a clutter is challenging due to the occlusion of the target caused by other objects, which leads to uncertainty. 
This problem occurs inevitably in many robotic manipulation settings, where the robot needs to adjust the viewpoint of its sensor to reduce grasping difficulty, i.e. from where the target object is less obscured in view, to plan a more successful grasp.

There have been multiple approaches to tackling such grasping tasks with a calibrated camera mounted on the end-effector of robot arm. 
A feasible solution is visual servoing~\cite{vs-1, vs-2} that perform continuous feedback with policy search methods~\cite{levine2016end} on visual features perceived by pose estimators~\cite{multiview-pose, multiview-1}. 
Nevertheless, vision systems constructed in such a way often rely on a number of hand-engineered components, and estimation of positions and poses are prone to errors for partially occluded objects. 
An alternative approach is a modular pipeline that decomposes vision and control into separate network modules~\cite{yan2017sim}. 
However, given undesirable recognition under occlusion condition, to make a reasonable decision is difficult.


\begin{figure}[t]
    \centering
    \includegraphics[width=0.98\linewidth]{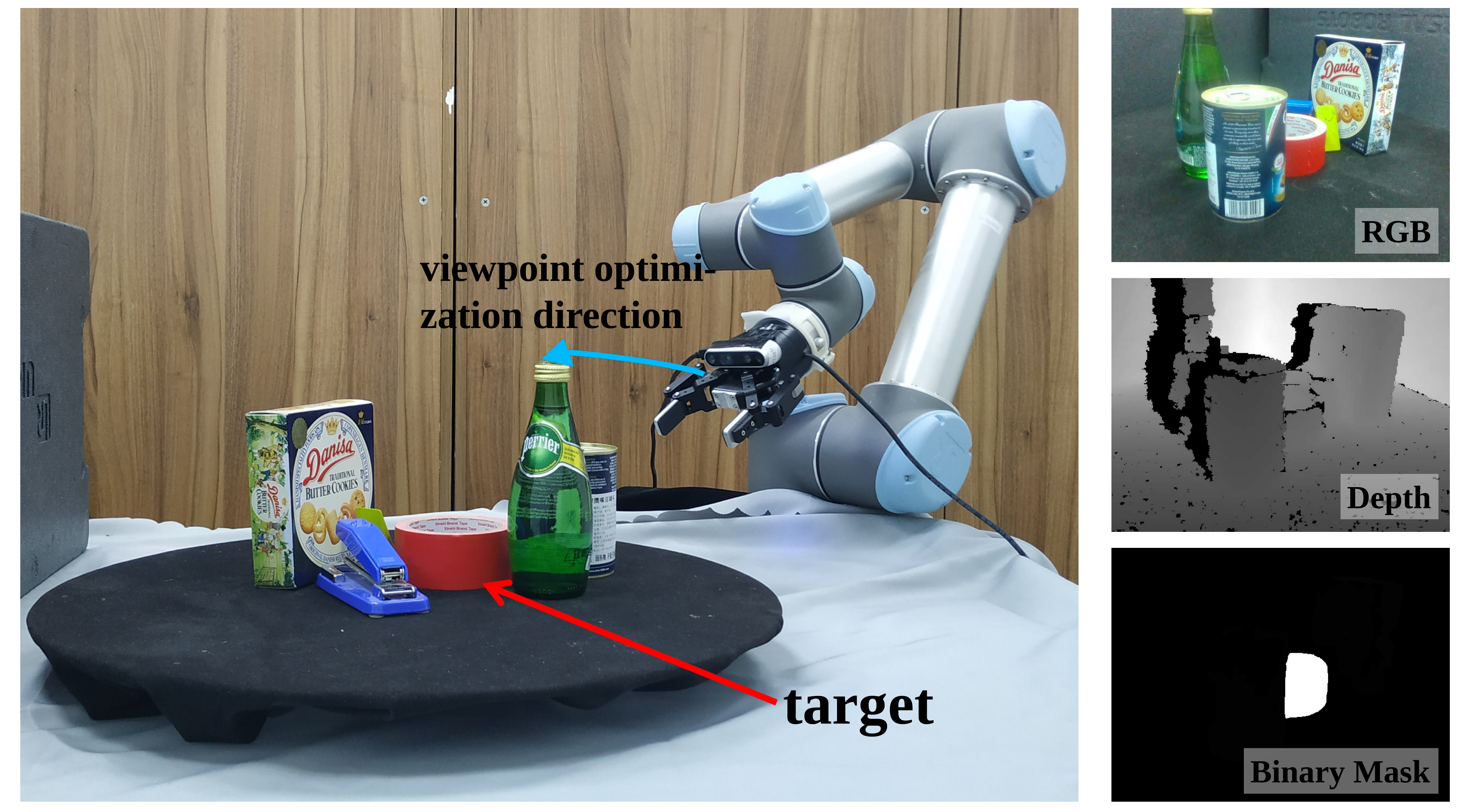}
    \caption{An example of our problem setting. The robotic arm is equipped with a RGBD camera and a gripper on its end-effector. The objective is to perform better grasping in clutter scenes by optimizing the viewpoint.}
    \vspace{-1\baselineskip}
    \label{fig:reality}
\end{figure}

In recent years, end-to-end deep reinforcement learning (RL) have proven successful in several challenging robotic manipulation tasks, including viewpoint optimization~\cite{calli2018viewpoint} with model-based~\cite{model-based} and model-free~\cite{model-free} techniques.
However, low sample efficiency~\cite{gu2016q}, heavy expenses, low reproducibility, and safety issues remain major problems for real-world training, while the gap between simulation and reality can be a significant challenge. In addition, the transferability of RL~\cite{gamrian2018transfer} poses another challenge: the performance of RL algorithms can drop dramatically when meeting an unseen scene.

Our work to address these challenges is inspired by embodied AI systems~\cite{savva2019habitat}. 
By training embodied AI agents (virtual robots) in a simulated 3D world, the learned skills such as active perception and long-term planning can be transferred to reality and work immediately. 
However, these embodied AIs rely on photo-realistic simulators to learn, which is not practical for many real-world problems. 
We propose a new ``Real Embodied Dataset'' (RED) to largely reduce the gap between simulator and real-world. 
Specifically, in RED viewpoints are densely sampled from the upper hemisphere of the clutter, which enables us to virtually command the gripper to change its viewpoint and receive real visual feedback from dataset. 
The dataset contains 31K aligned high-quality RGB-D images collected on 173 clutter scenes of 17 household objects, with corresponding camera poses relative to the clutter and hand-annotated amodal segmentation masks indicating how occlusion is formed. 
With the occlusion information, RED can benchmark not only visual reasoning, but also decision making, in an inexpensive, safe, and reproducible manner.



Based on our new RED dataset, we develop an active grasping methodology that is capable of reasoning about 3D geometric data such as point clouds in a scene transferable manner. 
Our visual inputs are aligned RGB-D images captured by the end-effector camera (eye-in-hand view). 
We model active grasping as a 3-stage pipeline: 
1) an object detector module that extracts pixels of the grasping target as a binary segmentation mask on the point cloud;
2) a viewpoint optimizer that decides in which direction the camera should move, or if the current viewpoint is good enough to plan a grasp;
3) a grasp planner that proposes feasible grasping points and operates the robot.
We adopt existing work for object detection and grasp planning.
For viewpoint optimization, we design a novel and practical mask-guided reward function and augmented the PointNet architecture to achieve efficient policy learning with deep RL. 
The pipeline modules are logically independent from each other, which allows separate training and facilitates easy debugging and replacement and they also provide interpretable intermediate results for sanity check.


To ensure enhanced transferability, we incorporate many new insights. 
First, our reward design represents the quality of viewpoints with the ratio of visible pixels, which is modelled independent with regard to environments and object categories. 
Second, by taking 3D reasoning, our RL solver is able to learn common occlusion patterns after a large amount of viewpoint adjustments in the training phase. 
Therefore, our active grasping system is \emph{transferable}: given a new object detector, we can directly integrate it into our framework without re-training any module.

We evaluate our method and its variants in both simulation and reality with a 6-DOF UR-5 robotic arm, on the end-effector of which an Inter RealSense D435 depth camera is mounted (Fig. \ref{fig:reality}).
Experimental evaluation demonstrates that our transferable active grasping pipeline trained on our Real Embodied Dataset can achieve a 82.0\% success rate grasping in clutter scenes on a wide range of objects in reality without further training. 
The dataset and code are available at \href{https://github.com/cxy1997/Transferable-Active-Grasping}{\color{blue}{https://github.com/cxy1997/Transferable-Active-Grasping}}.

In summary, trying to tackle scene occlusion, our major contributions in this work include:
1) A large-scale multi-view Real Embodied Dataset to enable real visual feedback;
2) A 3-stage active grasping pipeline to support both model transferability and interpretability; 
3) A mask-guided reward design to facilitate efficient training and robust grasping.

\section{Related Work}
\subsection{RGB-D Datasets}

There have been multiple datasets studying semantic segmentation and pose estimation on RGB-D images~\cite{segd-1, segd-2, segd-3}.
However, most of them either consist of synthetic scenarios \cite{hand-data-1, hand-data-2}, or contain limited scenes \cite{small-data-1, small-data-2}.
Recently, a few further study occlusion information~\cite{occd-1, occd-2}, while they only perform detection or estimation from limited viewpoints.
None of existing datasets provide large scale multi-view real-world RGB-D data of occlusion reasoning in clutter scenes.


\subsection{Object Detection}
Object detection techniques have been a powerful tool for scene understanding that can operate on images~\cite{redmon2016you, girshick2015fast, ren2015faster, wang2019pseudo} and pointclouds~\cite{li2018pointcnn, yang2018pixor}.
However, these methods can hardly handle object occlusion in clutter scenes, which may mislead the action of a robot.
We follow the amodal semantic segmentation setting\cite{zhu2017semantic, li2016amodal} to help the robot perceive the target more reasonably.

\subsection{Reinforcement Learning}
RL has shown promising results in a variety of domains in recent years~\cite{rl-app1, rl-app2, rl-app3}, with robotics being its most direct application~\cite{gu2017deep, levine2016end, wang2019tendencyrl}. However, RL methods often suffer from sample efficiency and scalability problems, which is particularly serious in robotic manipulation~\cite{sample-efficiency}. 
Off-policy RL methods~\cite{levine2018learning} have shown improvements in this direction, but they still require a large amount of time to collect data. 
Another approach is to plan grasping via a hierarchical of low-level policies (HRL) \cite{hrl}, which has certain scalability but depends on manually defined low-level policies. 
Our method integrates the amodal segmentation information into RL, solving the sample efficiency problem and guaranteeing transferability.

\subsection{Imitation Learning}

Recently, imitation learning (IL) is widely applied in various domains~\cite{il-ap1, il-ap2}, especially robotic manipulations~\cite{il-app1, il-app2}. 
Given human demonstration data, the agent learns a policy that fits the behaviors of experts~\cite{il}. IL has good performances in some simple cases~\cite{il-1, il-2}. Nevertheless, as the task becomes more complex, the demand for demonstration data will also increase significantly. 
Besides, demonstration data is often designed task-specific and thus hard to be transferred, which is unrealistic for real applications. Some works try to solve this problem with hierarchical learning structure~\cite{il-hil}. However, it is difficult to plan in the long term while organizing the primitive skills.

\subsection{Grasping}
Grasping is an important primitive in robot manipulations, with target localization being its first and most important step. There exist many methods to plan long-term grasps, like motion planning~\cite{motion-planning-1} and pose estimations \cite{multiview-pose}. The detection process becomes relatively difficult in occluded environments, some work~\cite{petrelli2011repeatability, papazov2010efficient} tend to directly recognize the target even if it is partially visible. Others apply multi-view systems to this problem~\cite{multiview-pose}, in which case the performance depends on the selection of view points. Our approach simplifies the detection process of grasping and make it trainable via reinforcement learning (RL) with a simple and reasonable reward design.

\section{Real Embodied Dataset}

\begin{figure}[htbp]
    \centering
    \vspace{-0.5\baselineskip}
    \includegraphics[width=0.98\linewidth]{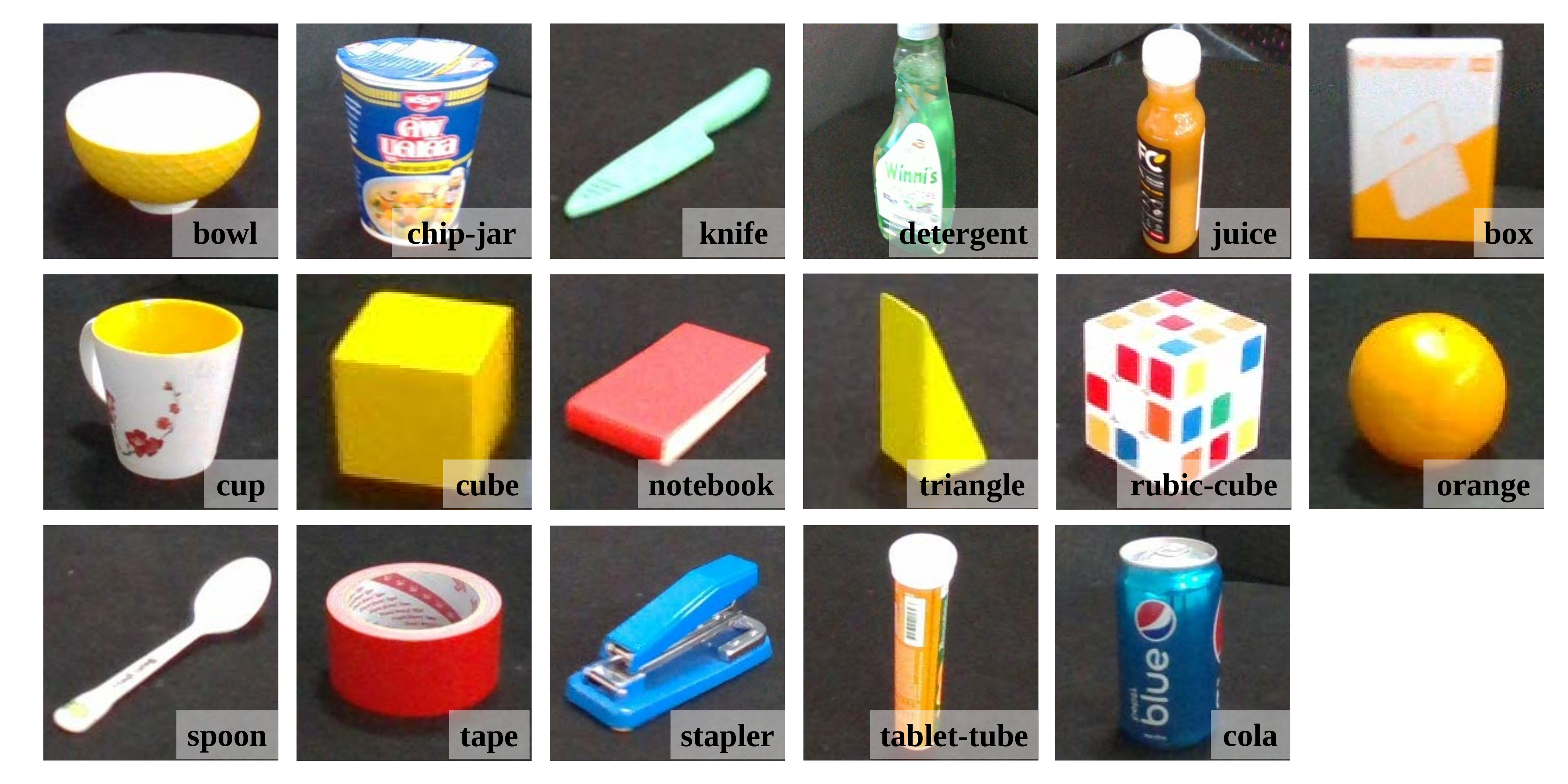}
    \vspace{-0.5\baselineskip}
    \caption{Object classes in our Real Embodied Dataset. Each object shown above represents a different category.}
    \label{fig:objects}
\end{figure}

\vspace{-0.5\baselineskip}
To efficiently train network modules of the active grasping system and to bridge the gap between simulation and reality in a most straightforward way, we constructed Real Embodied Dataset (RED), a large-scale multi-view dataset consisting of more than 31K RGB-D images collected from 173 clutter scenes.
RED is specifically intended as an embodied AI platform for occlusion recognition and policy learning. Its direct use of real-world images help to improve model transferability compared synthetic data obtained from photo-realistic simulators.

\subsection{Object Selection}

The dataset includes a preselected set of 17 common household items, which differ in size, shape, color texture, transparency and would pose various levels of occlusion.
Fig. \ref{fig:objects} shows the full set of items.
Tall, thin objects, such as a bottle, are likely to form partial occlusion with other regular-shape objects.
Non-convex objects, such as a bowl, are more difficult to detect and properly grasp as structured light would form blank zones in their depth images.

\subsection{Dataset Design}

\begin{figure}[htbp]
    \vspace{-0.5\baselineskip}
    \centering
    \includegraphics[width=0.15\linewidth]{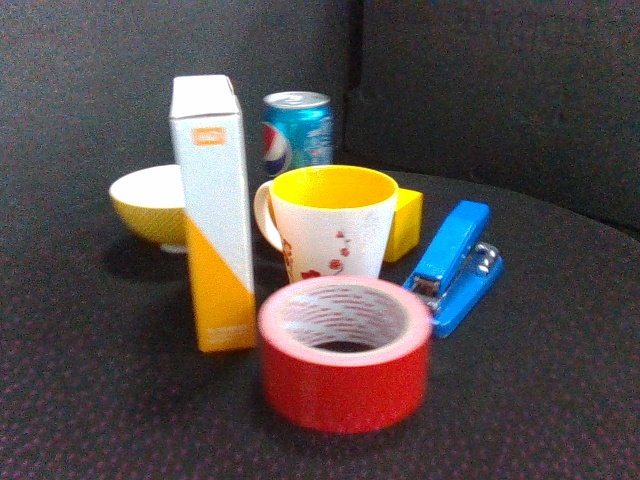}\hspace{-0.4em}
    \includegraphics[width=0.15\linewidth]{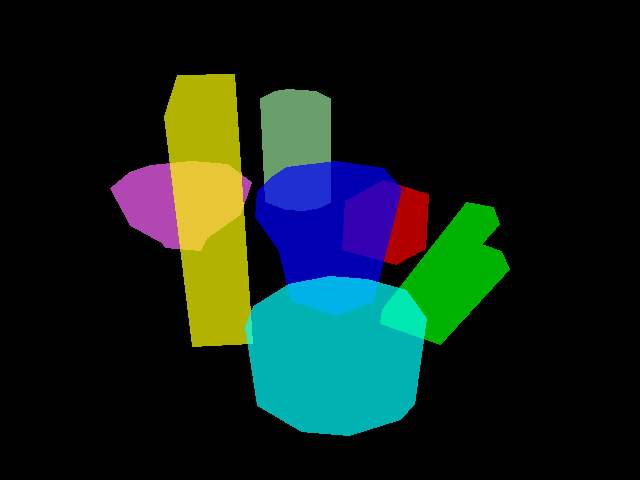}
    \includegraphics[width=0.15\linewidth]{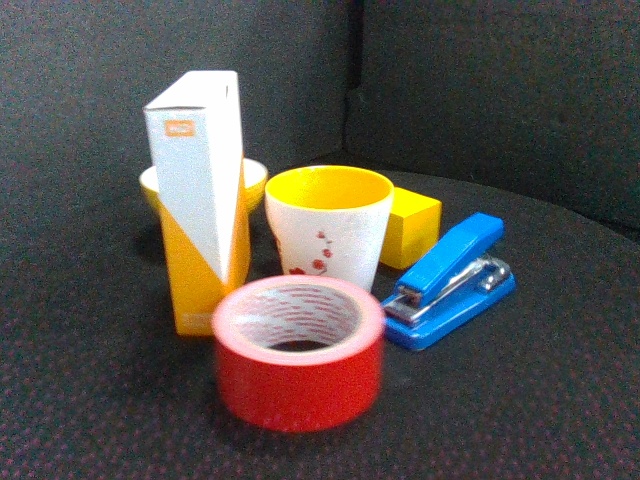}\hspace{-0.4em}
    \includegraphics[width=0.15\linewidth]{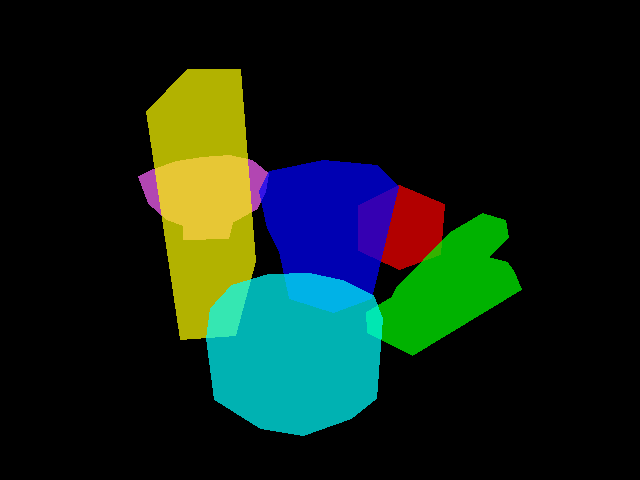}
    \includegraphics[width=0.15\linewidth]{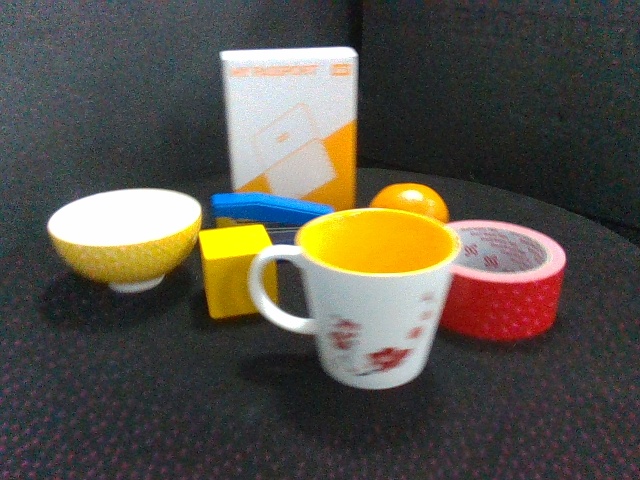}\hspace{-0.4em}
    \includegraphics[width=0.15\linewidth]{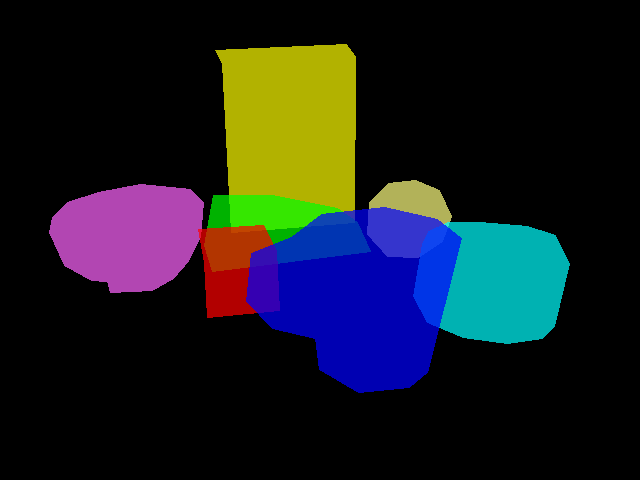}
    
    \vspace{0.2\baselineskip}

    \includegraphics[width=0.15\linewidth]{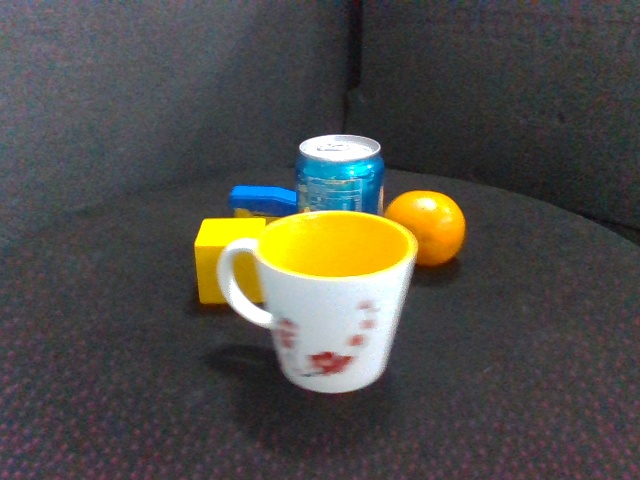}\hspace{-0.4em}
    \includegraphics[width=0.15\linewidth]{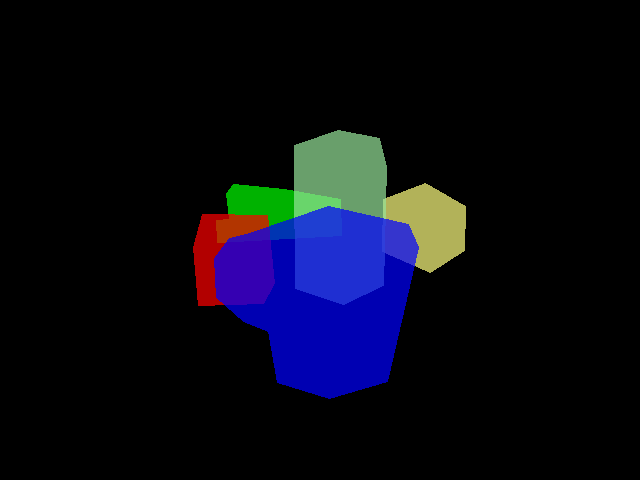}
    \includegraphics[width=0.15\linewidth]{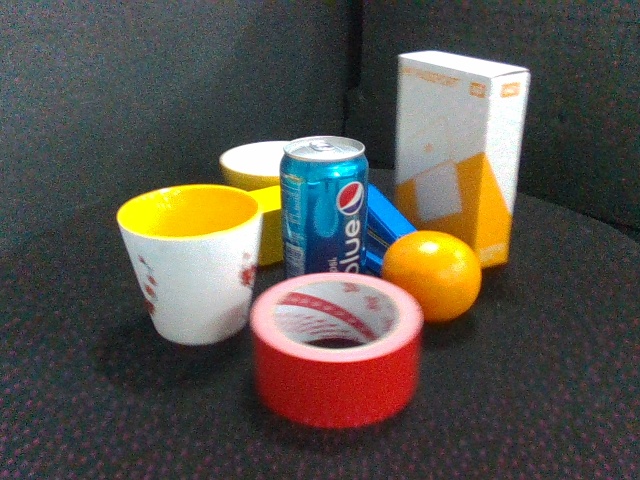}\hspace{-0.4em}
    \includegraphics[width=0.15\linewidth]{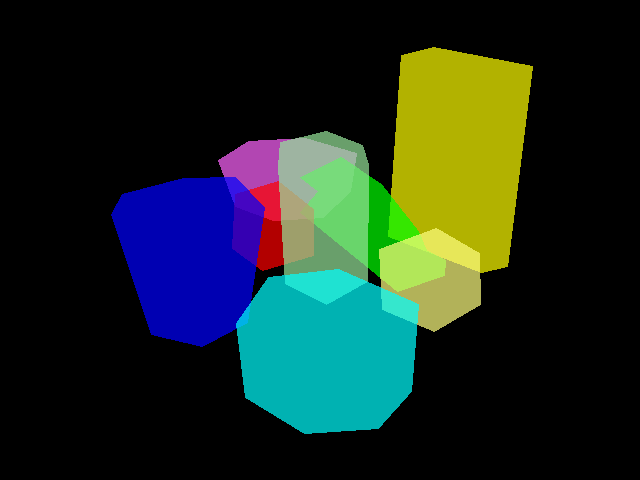}
    \includegraphics[width=0.15\linewidth]{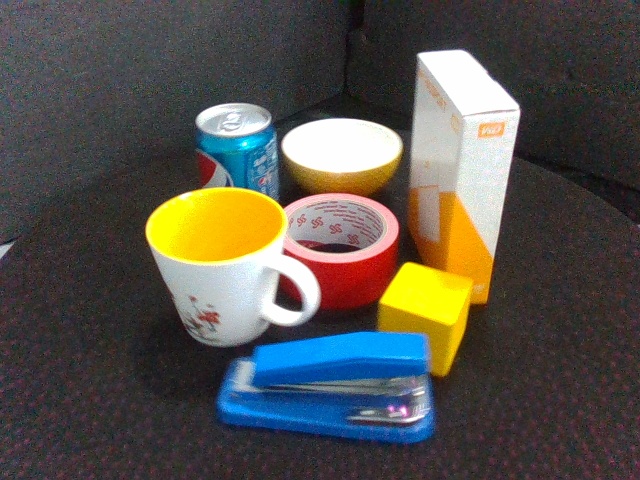}\hspace{-0.4em}
    \includegraphics[width=0.15\linewidth]{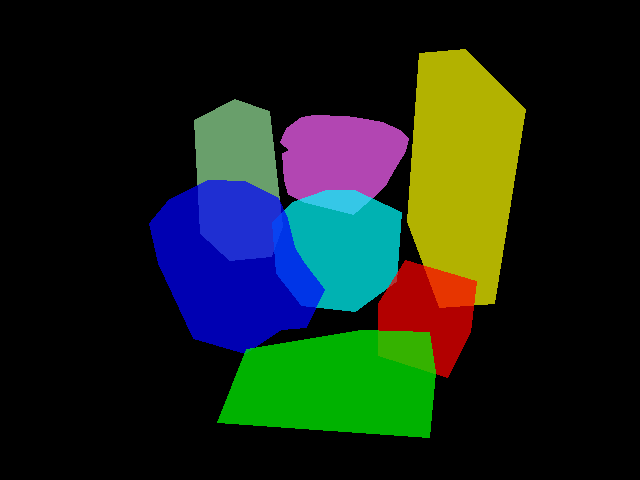}
    
    \vspace{0.2\baselineskip}

    \includegraphics[width=0.15\linewidth]{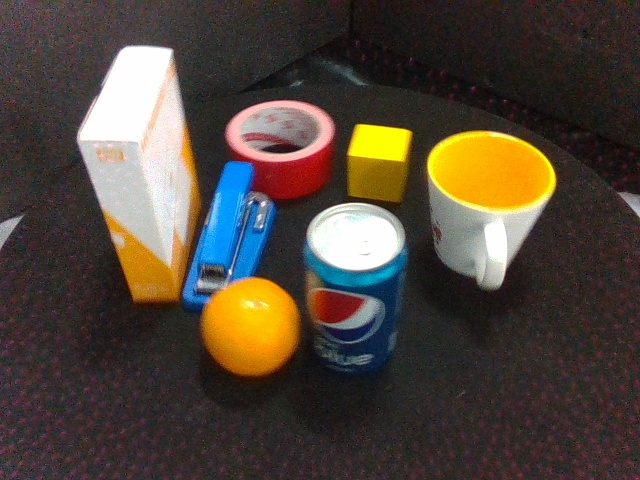}\hspace{-0.4em}
    \includegraphics[width=0.15\linewidth]{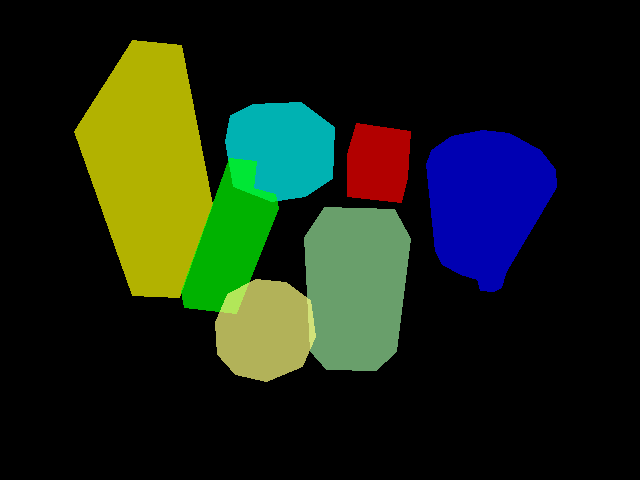}
    \includegraphics[width=0.15\linewidth]{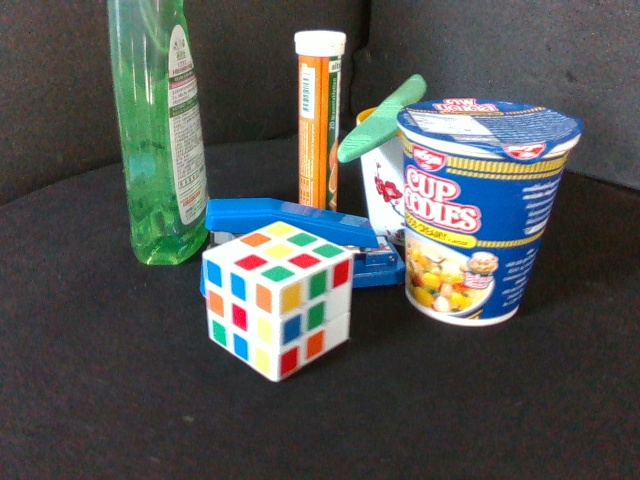}\hspace{-0.4em}
    \includegraphics[width=0.15\linewidth]{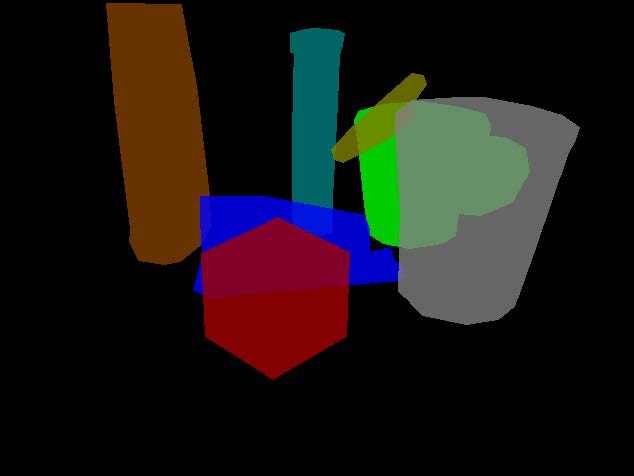}
    \includegraphics[width=0.15\linewidth]{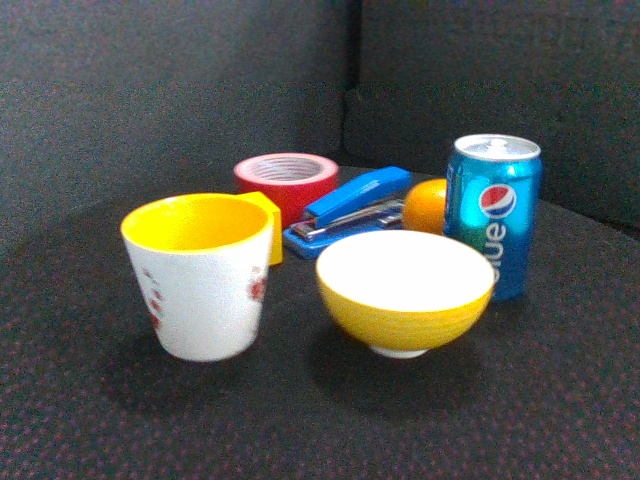}\hspace{-0.4em}
    \includegraphics[width=0.15\linewidth]{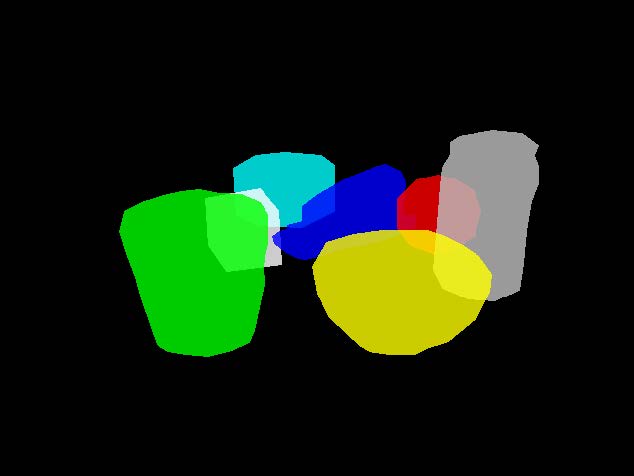}

    \vspace{-0.5\baselineskip}
    \caption{Examples of semantic amodal segmentation annotation in the Real Embodied Dataset. For each RGB-D image, complete segmentation masks of objects are provided, along with their order of occlusion. }
    \vspace{-0.5\baselineskip}
    \label{fig:clutters}
\end{figure}

The intention of the dataset is to provide a large scale multi-view representation of object occlusion in clutter scenes, which assists the learning of object detection and segmentation, occlusion inference and other embodied AI applications. Fig. \ref{fig:clutters} shows some of the clutter settings with corresponding annotations.

For each clutter scene, RED provides 180 aligned high-quality RGB-D images with corresponding camera poses, sampling the viewing hemisphere.
With known camera intrinsics, these RGB-D images can be unprojected as corresponding pointclouds.
Specially, the dataset provides amodal instance segmentation annotation for all images.
That is, we annotate full segmentation masks including occluded parts for all objects in an image, as well as their occlusion order.
In this way, occlusion can be reproduced by placing object segmentation masks on top of each other in that order.

With annotations described above, the dataset not only contains sufficient information for regular object detection and segmentation learning, it is also helpful for inferring complete object shapes from their visible parts.
To the best of our knowledge, RED is the first dataset to study object occlusion on a large scale using amodal segmentation masks from a multi-view perspective.

\subsection{Data Collection}

The dataset is collected using an Intel RealSense Depth Camera D435~\cite{realsense} with known camera intrinsics mounted on the end-effector of a UR-5 robotic arm, which captures aligned, synchronous RGB-D images at a $480 \times 640$ resolution.

\begin{figure}[t]
    \centering
    \includegraphics[width=0.5\linewidth]{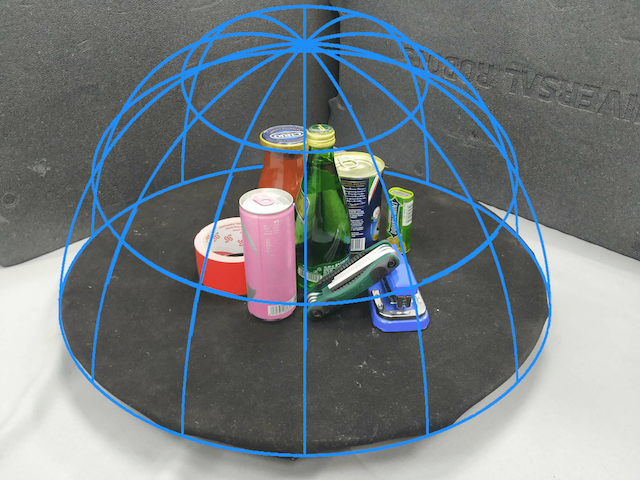}
    \caption{The 180 viewpoints are sampled from the upper hemisphere of the clutter, at an interval of 10 degrees of longitude and latitude. }
    \vspace{-0.7\baselineskip}
    \vspace{-1\baselineskip}
    \label{fig:viewpoints}
\end{figure}

During data collection, objects in clutter are placed on a turnable spinning at a constant angular velocity of $0.25\pi\ rad/s$.
The RGB-D sensor collects data at 4.5Hz, thus distributing viewpoints at an interval of 10 degrees.
In addition, the robotic arm moves its end-effector along a quarter-circular arc with a radius of 0.5m centered at the turnable center, and collects data at $30^{\circ}$, $40^{\circ}$, $50^{\circ}$, $60^{\circ}$ and $70^{\circ}$ above the horizon of turnable.
In total, RGB-D images from 180 uniformly distributed viewpoints are collected for each clutter scene.
Fig. \ref{fig:viewpoints} shows the distribution of viewpoints on the upper hemisphere of clutter.

To acquire accurate amodal instance segmentation masks, each clutter scene is scanned multiple times from the same set pf viewpoints.
Each time, a new object is added to the clutter, whose \emph{complete} segmentation masks are then annotated.
Finally, after all objects have been added to a clutter scene, occlusion orders are labelled on the collected images.

\subsection{Real Embodied Simulator}
With all aforementioned data at hand, it is quite natural to derive a real-data simulator directly from the dataset.
With a given robot pose, to form a genuine feedback, one only needs to match it with the closest data point stored in the dataset and add noises if necessary. 
Our real embodied simulator is completely data-driven.
It is ecological in computational cost, and can achieve a throughput of thousands of frames per second without interfering with model training. 
Also, since our real embodied simulator does not involve any rendering, its feedback to the embodied AI agent is 100\% genuine, and the trained agent will be robust enough to transfer to real-world scenarios and perform well without further finetuning.
\section{Transferable Active Grasping Framework}
We consider the problem of planning a robust parallel-jaw grasp for a targeting object in a clutter scene with an eye-in-hand camera.
We propose a transferable active grasping framework that decomposes the grasping task into three stages: (i) object detection, (ii) viewpoint optimization, and (iii) grasp planning. 
The main novelty lies in the viewpoint optimization module, which enables the learned policy to generalize across different object layouts and transfer between simulation and reality. 
The modular design allows for the model to produce understandable intermediate results, and to adapt flexibly to new scenes with a large proportion of parameters unchanged.

\begin{figure*}[h]
    \centering
	\includegraphics[width=0.9\linewidth,height=9\baselineskip]{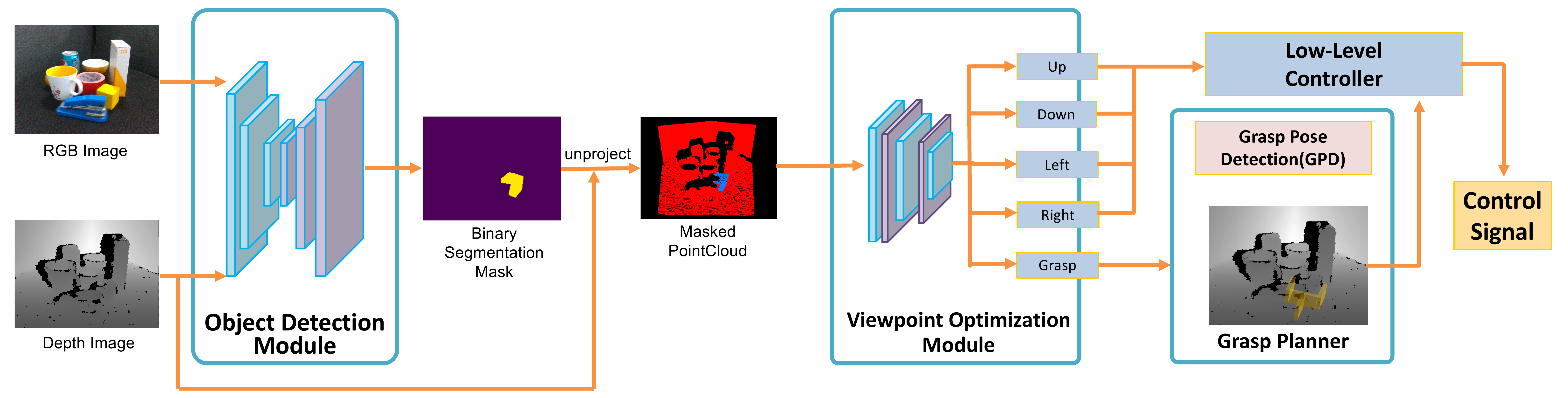}
    \caption{Architecture of our proposed transferable active grasping framework. The modular pipeline is composed of an object detector, a viewpoint optimizer, and a grasp planner. Interfaces between modules are binary segmentation masks and high-level commands, which can be easily understood by human.}
    \vspace{-1.5\baselineskip}
    \label{fig:architecture}
\end{figure*}

The architecture of the transferable active grasping framework is shown in Fig. \ref{fig:architecture}.
At each timestep $t$, the model takes an RGB-D observation $o_t$ as input, and produces a control signal $u_t$.
The control signal outputted is either an adjustment in viewpoint or a grasping proposal, which will be transmitted by a low-level controller to torques of the robot's motors and end-effector gripper. We introduce all major sub-modules in the following sections. 


\subsection{Object Detection}
In general, the object detection module takes an RGB-D image as input and outputs a binary segmentation mask indicating whether each pixel belongs to the target object.
This is achieved by training a segmentation network that assigns a label to each pixel indicating targeting object or not. 
To produce a binary segmentation mask for viewpoint optimization and grasping at later stages, the pixels corresponding to the target class are labeled $1$, while all other pixels are labeled $0$.

There have been many approaches that serves this purpose from different perspectives, and we offer multiple implementations of the object segmentation module.
Semantic segmentation directly performs pixel-wise classification, taking advantage of rich feature representation extracted by deep backbone networks. 
Instance segmentation, on the other hand, detects all objects in an image and segments each instance, usually based on proposals. 
These classic segmentation methods are developed to handle 2D images, and can be extended to 3D tasks by adding an additional depth channel.
There are also pure 3D networks that operate directly on 3D coordinates to capture shape patterns.
Our implementation has covered all these choices for model comparison and selection, and new models can be incorporated easily into our flexible framework.

\subsection{Viewpoint Optimization}
\label{sec:vo}
Viewpoint optimization module is the core component of our active grasping framework. 
Instead of detecting and planning grasps from a fixed viewpoint, optimizing viewpoints allows the robot to form better detection and grasp proposals. 
As discussed in related work, the use of imitation learning is hindered by its requirement for massive demonstration samples. 
We instead adopt a deep reinforcement learning framework, which models the task setting as a Markov decision process (MDP): At each time step $t$, the robot observes the state as $s_t$, executes an action $u_t \in \mathcal{A}$ according to its policy $\pi$, and receives a reward $r_t$. 
The grasping policy is then optimized by maximizing its expected cumulative rewards, also known as the \emph{state value function} $V$ that represents how good is a state for an agent to be in:
\begin{equation}
	V^{*}(s_t) = \max_{\pi} \mathbb{E}\left[\sum_{k=0}^{\infty} {\gamma}^k r_{t+k} \mid \pi \right]
	\label{eqn:value}
\end{equation}


To obtain a general viewpoint optimization policy, we build the RL training environment completely from real-world data. 
Such data-driven simulation can lead to a great improvement of training efficiency, as real-world training would take weeks to finish, despite of safety concerns. 
Also, there is undoubtedly no gap between observations in real embodied simulation and reality, and therefore no need for domain transfer or fine-tuning.
The environment settings are described as follows:

\paragraph{Observation} The input observation of the viewpoint optimization module is a pointcloud (frustum) with a binary mask indicating the target. 
Such representations can largely reduce domain gaps by removing all unnecessary color / texture information. 

\paragraph{Action Space} At each timestep, the end-effector camera can move along a meridian or parallel on the viewing hemisphere by 10 degrees, or decide to plan a grasp from its current position: 
$\mathcal{A} = \{L, R, U, D, grasp\}$, where $L$, $R$, $U$, and $D$ are action of moving left, right, up and down the hemisphere, respectively.

We use the advantage actor critic policy gradient (A2C) algorithm to the viewpoint optimizer.
To achieve robust feature learning on unordered pointclouds, our actor-critic architecture adopts PointNet as its backbone network to extract shape features, and uses two MLPs to approximate the value function and Q-function, respectively.

\subsection{Mask-guided Reward}

Reward shaping has been critical for deploying RL algorithms. 
In active grasping, there two major considerations: 
First, the reward function should be general across different objects and environments. 
Second, sparse rewards should be avoided to facilitate efficient policy learning. 
Our mask-guided reward can successfully address these two challenges using the differential of object visibility.

\paragraph{Visibility} To quantitatively measure the occlusion level of objects, we define visibility $\psi$ for an object $c$ from viewpoint $v$ as
\begin{equation}
	\psi(c, v) = \frac{\left|\text{visible region of }c \text{ observing from } v\right|}{\left|\text{complete shape mask of }c \text{ observing from } v\right|},
\end{equation}
where $\left|\cdot\right|$ denotes the area of a segmentation mask.
Since the precise hidden area or volume is hard to measure, it is estimated by the number of pixels in the mask.

\paragraph{Grasping Reward} Ideally, the goal of viewpoint optimization is to navigate the end-effector camera to an optimal viewpoint $v^*$ for target object $c_{target}$, where a grasp is then attempted.
However, finding exactly the best viewpoint is unnecessary and difficult for a grasping task. Thus, to avoid unnecessary viewpoint refinement, we relax the goal to finding a viewpoint $v$ that satisfies
\begin{equation}
	\psi(c_{target}, v) > \delta,
\end{equation}
where $\delta$ is a visibility threshold determined by detection and grasping performance. We further introduce a grasping reward specially designed for the $grasp$ action:
\begin{equation}
r_t^{(grasping)}=
\begin{cases}
0.25, & \text{$a_t = grasp \land \psi(s_t) > \delta$}\\
\psi(s_t)-1.5, & \text{$a_t = grasp \land \psi(s_t) \leq \delta$}\\
0, & \text{otherwise}
\end{cases}
\label{eqn:reward_g}
\end{equation} 


\paragraph{Tendency Reward} Grasping reward alone is too sparse for RL training. 
Following~\cite{wang2019tendencyrl}, we introduce the tendency reward to measure whether an action has the tendency of moving to a better viewpoint:
\begin{equation}
\begin{split}
	r_t^{(tendency)} &= \psi(s_{t+1}) - \psi(s_t)\\
	         &= \psi(c_{target}, v_{t+1}) - \psi(c_{target}, v_t)
\end{split}
\label{eqn:reward_t}
\end{equation}
In other words, $r_t^{(tendency)}$ represents the improvement of target object visibility obtained in timestep $t$. 
By providing dense guidance signals at all timesteps, training efficiency can be boosted with the tendency reward. 
 
\paragraph{Final Reward}
The final reward function for RL training is a combination of grasping and tendency rewards:
\begin{equation}
	r_t = r_t^{(tendency)} + r_t^{(grasping)}
\end{equation}

\subsection{Grasp Planning}

The grasp planning module is invoked when the viewpoint optimization module decides to take the $grasp$ action.
Grasps are planned by detecting eligible grasping points on the input point cloud.
The proposed grasping points are converted to end-effector coordinates of the robot and then fed to the robotic control module.

The grasp pose detection (GPD) algorithm~\cite{gpd-1} is the core component of the grasp planner. 
GPD takes a viewpoint cloud with an ROI as input. 
It samples a large set of grasp candidates $\{H_i\}$ and concatenates their surface normals to image pixels.
A CNN is trained to classify these candidates. 
We train a LeNet~\cite{lenet} using data generated from BigBIRD dataset~\cite{bigbird}. 
Grasp labels are created by evaluating whether they are frictionless antipodal grasps. 
The network assigns each candidate $H_i$ a score $s_i \in [0,1]$, which indicates the probability for a candidate to be a valid grasp.

\subsection{Robotic Control}

The robotic control module continuously generates torque control signals for robot motors from the high-level actions and grasp plans.
The module processes high-level control signals in two steps.
Initially, target end-effector positions produced by previous stages in the pipeline are converted to joint positions of the robotic arm based on inverse kinematics.
The computed joint positions are then fed as targets to a PID controller, which adjusts the torques of the robot's 6 motors at high frequency to reach its current target joint position. 
Machine learning methods are not used in this module.
\section{Experiments}
Experiments are divided into three parts. First, we present and evaluate multiple choices of detectors on our dataset (Sec. \ref{sec:detector}). Second, we show that the combination of PointNet and RL outperforms its counterparts on viewpoint optimization (Sec. \ref{sec:viewpoint}). At last, we provide detailed experiments both in simulation and in real world to evaluate the whole framework with different implementations (Sec. \ref{sec:whole}).

\subsection{Object Detection Module}
\label{sec:detector}
In this section we implement and evaluate some state-of-the-art segmentation models on our Real Embodied Dataset.
We adopt WideResNet + DeepLabV3 (semantic segmentation), Mask-RCNN (instance segmentation), and PointNet for segmentation (point cloud segmentation), respectively, as implementations of the object detection module.
In order to reduce memory footprint, we additionally adopt In-Place Activated BatchNorm to double the training batch size for better performance.

To improve model robustness and to make image segmentation models comparable to point cloud segmentation methods in terms of input information, we further construct a few variants of segmentation models that utilizes depth images.
We follow the work of~\cite{ma2017multi} and represent depth information as additional input channel(s). We fuse depth with color by concatenating channels of raw pixel (pixel-concat) or on the feature level (feature-concat).

Our dataset is divided into 70\% training data, 15\% validation data and 15\% test data for training and evaluation. Model selection is done based on performance on the validation set, and the selected model is evaluated in the test dataset.

We compare performance of different implementations in Table \ref{tab:semantic} and \ref{tab:instance}.
All models presented are able to achieve desirable segmentation performance on the test set.
We also observe that there is no significant difference between models with different RGB-D fusion methods.

\begin{table}[H]
\centering
\vspace{1\baselineskip}
\begin{tabular}{lcccc}
\Xhline{2\arrayrulewidth}
Depth Fusion & \tabincell{c}{pixel\\acc.} & \tabincell{c}{mean\\acc.} & \tabincell{c}{mean\\IU} & \tabincell{c}{f.w.\\IU} \\
\hline
No-Depth & 98.52 & 95.02 & 90.89 & 97.16 \\
Pixel-Concat & 98.45 & 94.56 & 90.48 & 97.07 \\
Feature-Concat & 98.41 & 93.75 & 90.18 & 96.94 \\
\Xhline{2\arrayrulewidth}
\end{tabular}
\caption{Semantic segmentation accuracy on Real Embodied Dataset with wide-ResNet+Inplace-ABN.}
\vspace{-1\baselineskip}
\label{tab:semantic}
\end{table}

\vspace{-1\baselineskip}
\begin{table}[H]
\centering
\begin{tabular}{llccccccc}
\Xhline{2\arrayrulewidth}
Depth Fusion & AP & AP$_{50}$ & AP$_{75}$ \\
\hline
No-Depth & 93.17 & 99.82 & 99.82 \\
Pixel-Concat & 78.22 & 99.36 & 95.00 \\
Feature-Concat & 79.56 & 99.51 & 95.82 \\
\Xhline{2\arrayrulewidth}
\end{tabular}
\caption{Instance segmentation accuracy on Real Embodied Dataset with Mask R-CNN.}
\vspace{-2.5\baselineskip}
\label{tab:instance}
\end{table}


\subsection{Viewpoint Optimization Module}
\label{sec:viewpoint}
\begin{figure}[htbp]
    \centering
    \includegraphics[width=0.49\linewidth]{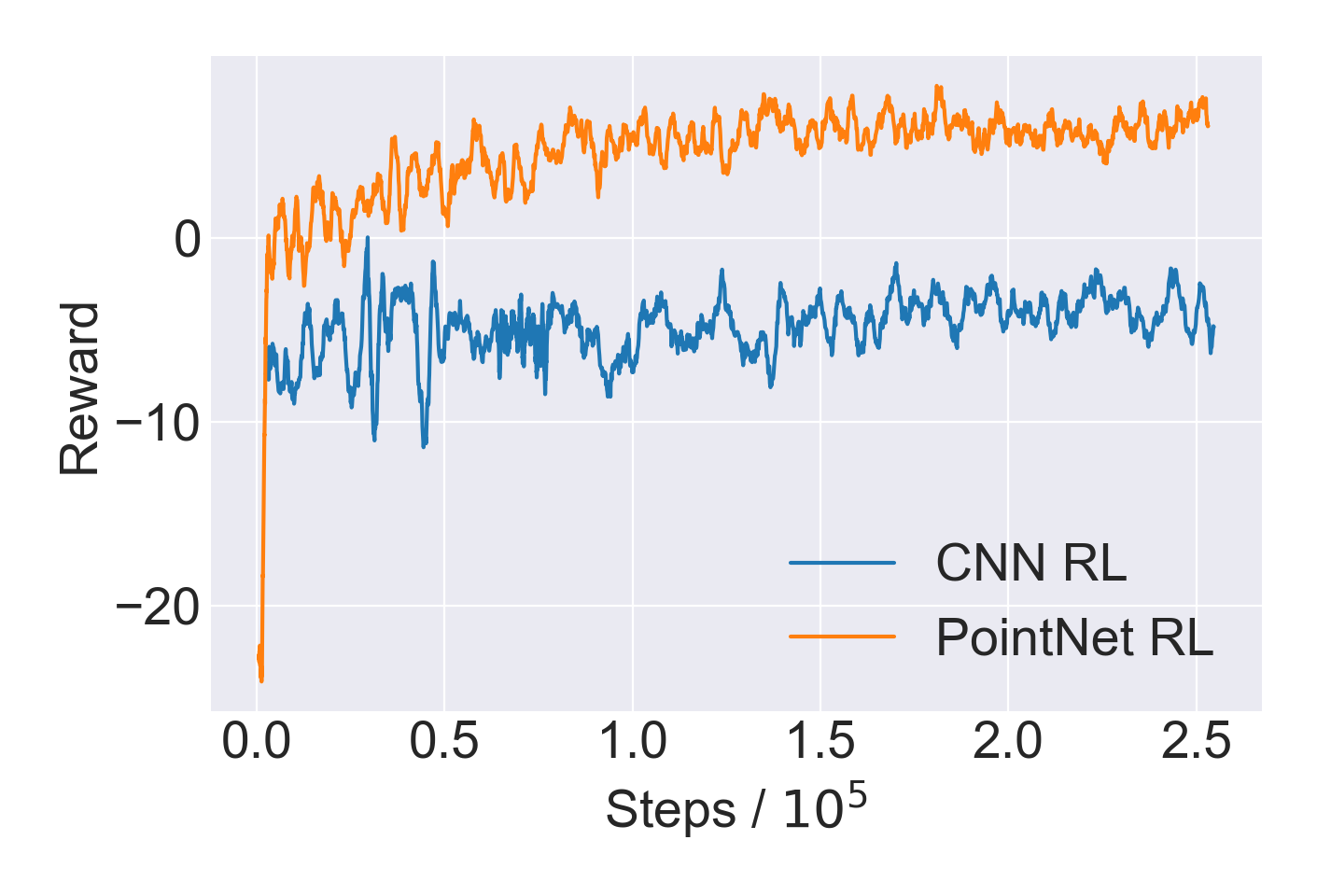}
    \includegraphics[width=0.49\linewidth]{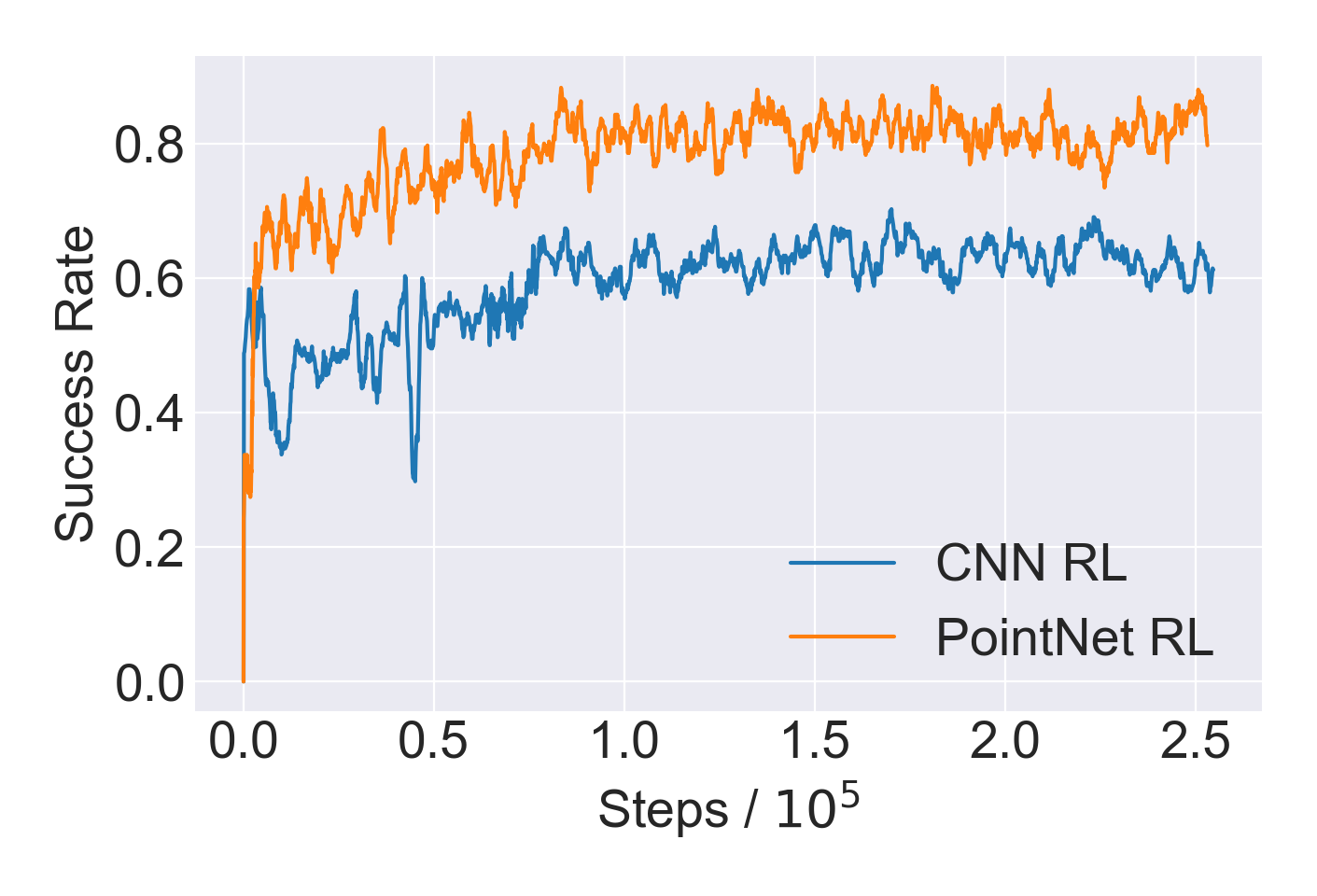}
    \includegraphics[width=0.49\linewidth]{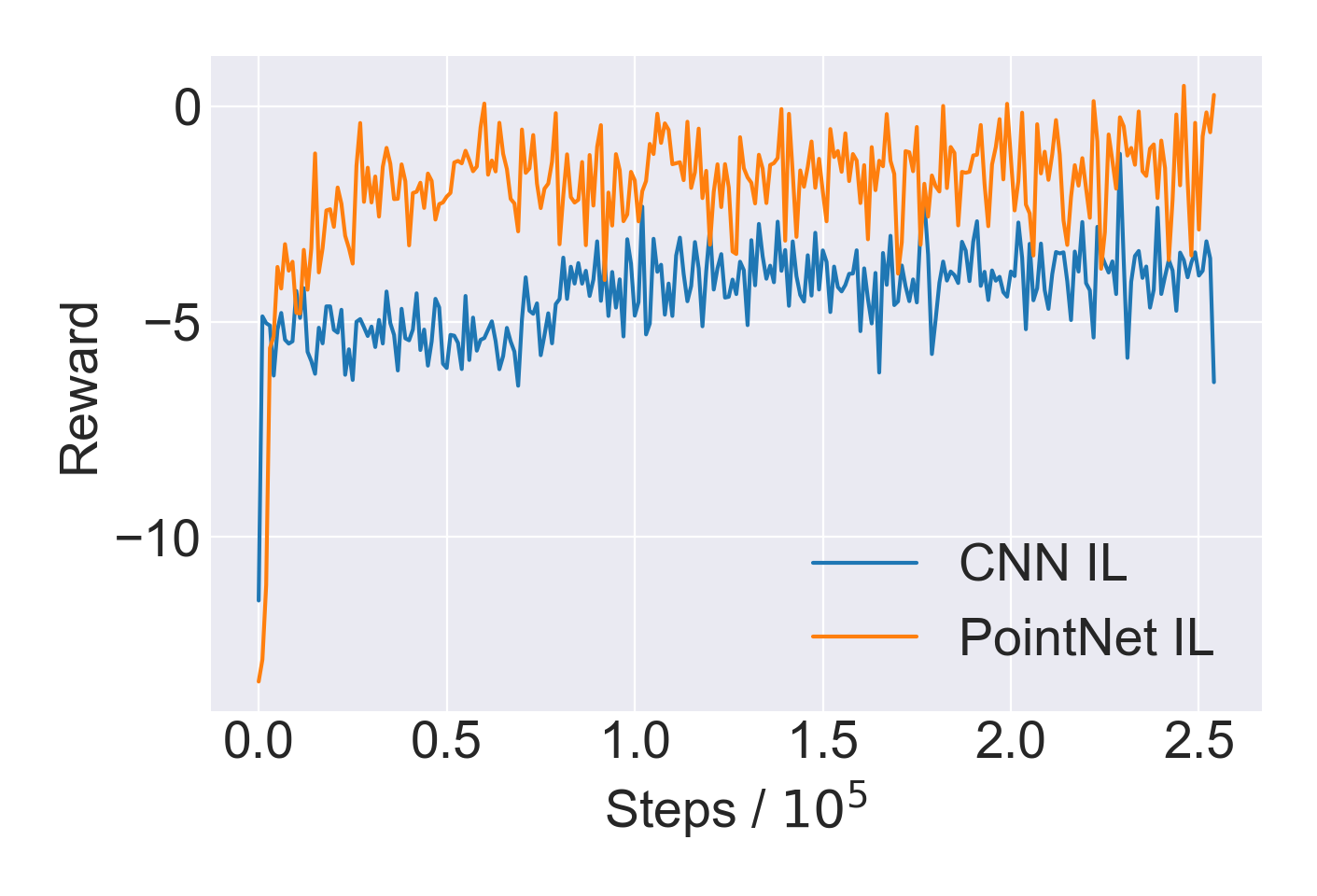}
    \includegraphics[width=0.49\linewidth]{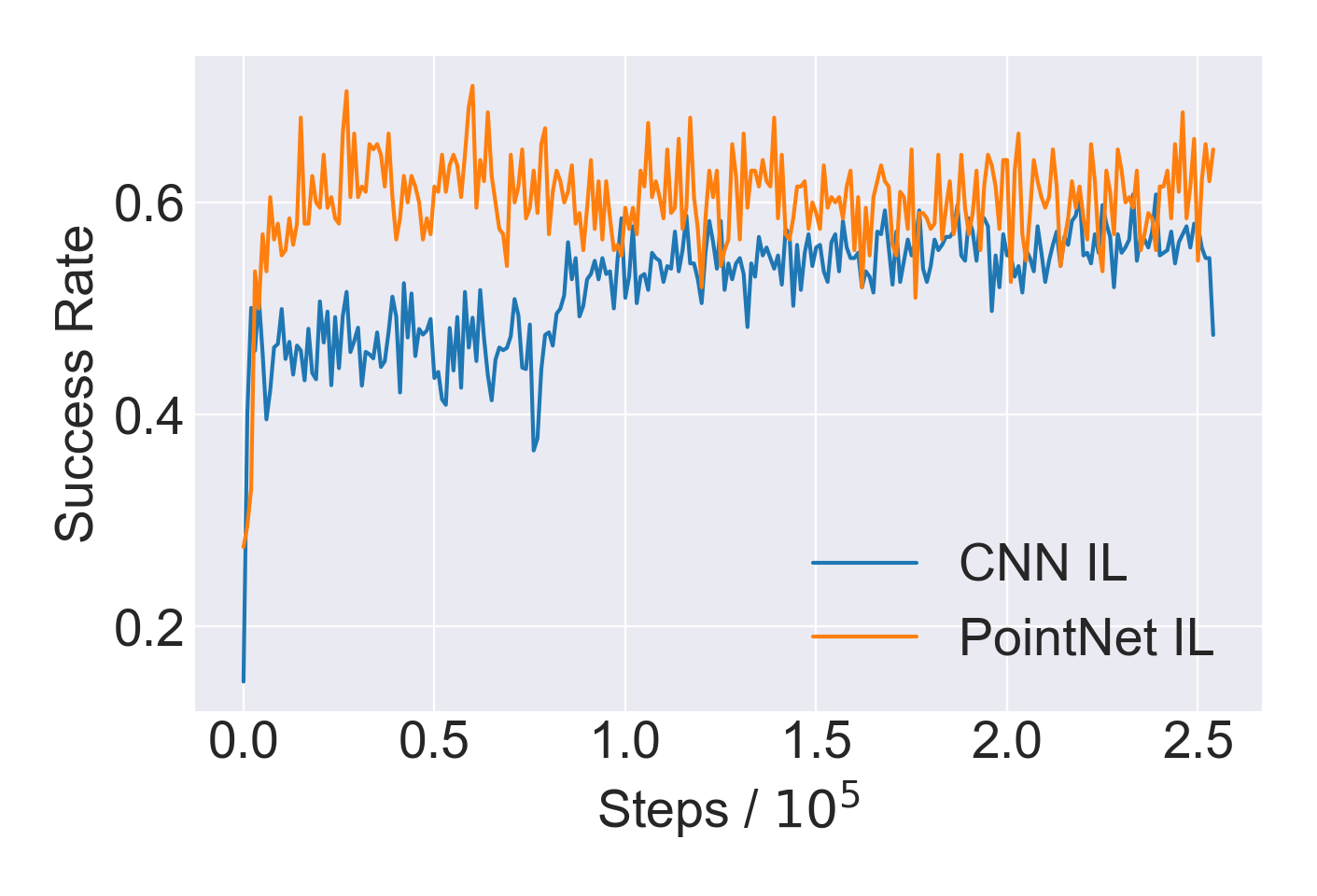}
    \vspace{-2\baselineskip}
    \caption{Learning curves of viewpoint optimization module against baselines. The plots are averaged over 5 runs with different random seeds. Grasping success is defined as perform grasping at a viewpoint with visibility of target object over threshold $\delta=0.9$.}
    \label{fig:curves}
\end{figure}

In this section, we inspect both network designs (PointNet \& 2D CNNs) and learning frameworks (RL \& 2D IL) for viewpoint optimization, in Fig. \ref{fig:curves}.
We replace the fully connected (fc) layers in PointNet with two fc layers with 512 and 256 cells as policy parameters.
The corresponding CNN model contains four convolutional layers consisting of 64 $10\times 10$, 64 $5\times 5$, 32 $3\times 3$ and 16 $3\times 3$ filters respectively, followed by three fc layers with 4096, 2048 and 1024 cells.
We use visibility threshold $\delta=0.9$ in grasping rewards.
The RL environment is constructed direct using ground truth data in the RED dataset, as is described in Sec. \ref{sec:vo}.
In IL, the model architecture is same as in RL, which is trained by supervised learning on 1,000 successful trajectories collected from reinforcement learning episodes.

Fig. \ref{fig:curves} shows that 1) reasoning on 3D spatial structure with PointNet significantly improves the learned policy; 2) reinforcement learning with our mask-guided reward shows better performance than imitation learning on the task of viewpoint optimization.
Since states distributed on a hemisphere can be reached from more than one route, it is natural for contradictory state-action pairs to co-exist in expert data.
This characteristic does not have influence on RL because it learns policies directly from rewards.
However, an imitation learning model fails to generate consistent action sequences when multiple actions are considered good.

\subsection{Transferable Active Grasping Framework Analysis}
\label{sec:whole}
We assemble the transferable active grasping framework and evaluate its success rate both in simulation and on a UR-5 robotic arm. Different implementations of sub-modules are compared in Table \ref{tab:sim} and \ref{tab:real}, with grasping success defined as perform grasping at a viewpoint with visibility of target object over threshold $\delta=0.9$.

The PointNet architecture shows a clear advantage over convolutional networks, and reinforcement learning on real-world data achieves better performance than imitation learning. 
Since GPD does not grarantee a 100\% grasping accuracy, the grasping success rate in reality is slightly lower than that in simulation, which still proves strong transferability of our framework.

\begin{table}[H]
\centering
\includegraphics[width=0.98\linewidth]{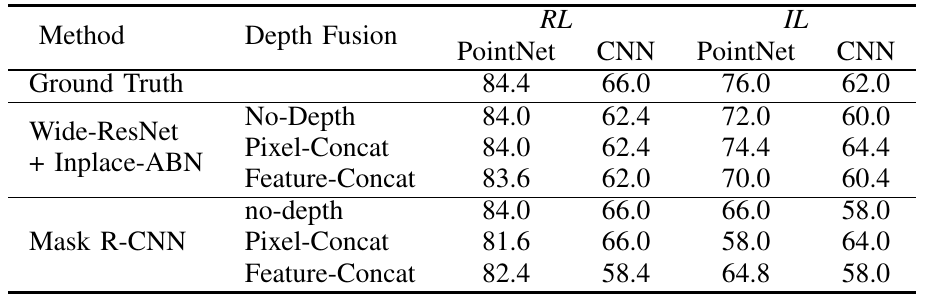}
\caption{Grasping success rate (\%) in simulation. }
\vspace{-3\baselineskip}
\label{tab:sim}
\end{table}

\begin{table}[H]
\centering
\includegraphics[width=0.98\linewidth]{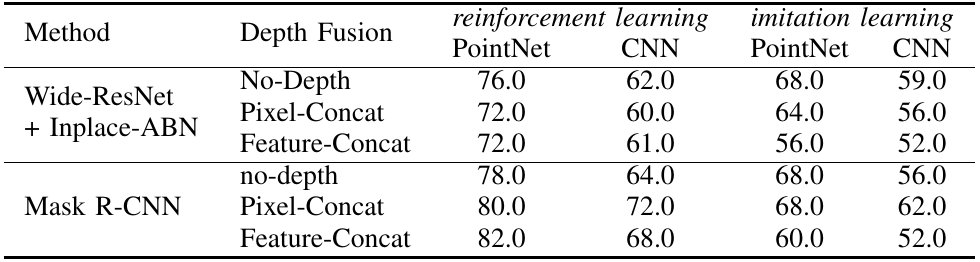}
\caption{Success rate of grasping on a real UR-5 robotic arm.}
\label{tab:real}
\end{table}
\vspace{-1\baselineskip}

\vspace{-1\baselineskip}
\section{Conclusion}
In this work, we proposed a large-scale multi-view Real Embodied Dataset (RED) as a data-driven simulator to facilitate photo-realistic training.
We developed a transferable active grasping pipeline that perform viewpoint optimization to achieve robust grasping in clutter environments.
In the future, it is worthwhile thinking of leveraging amodal segmentation in our pipeline.

\section{Acknowledgement}
This work is supported by Program of China, No. 2017YFA0700800, National Natural Science Foundation of China under Grants 61772332 and Shanghai Qi Zhi Institute.







\bibliographystyle{IEEEtran}
\bibliography{egbib}

\begin{thebibliography}{10}
\providecommand{\url}[1]{#1}
\csname url@samestyle\endcsname
\providecommand{\newblock}{\relax}
\providecommand{\bibinfo}[2]{#2}
\providecommand{\BIBentrySTDinterwordspacing}{\spaceskip=0pt\relax}
\providecommand{\BIBentryALTinterwordstretchfactor}{4}
\providecommand{\BIBentryALTinterwordspacing}{\spaceskip=\fontdimen2\font plus
\BIBentryALTinterwordstretchfactor\fontdimen3\font minus
  \fontdimen4\font\relax}
\providecommand{\BIBforeignlanguage}[2]{{%
\expandafter\ifx\csname l@#1\endcsname\relax
\typeout{** WARNING: IEEEtran.bst: No hyphenation pattern has been}%
\typeout{** loaded for the language `#1'. Using the pattern for}%
\typeout{** the default language instead.}%
\else
\language=\csname l@#1\endcsname
\fi
#2}}
\providecommand{\BIBdecl}{\relax}
\BIBdecl

\bibitem{vs-1}
N.~Vahrenkamp, S.~Wieland, P.~Azad, D.~Gonzalez, T.~Asfour, and R.~Dillmann,
  ``Visual servoing for humanoid grasping and manipulation tasks,'' in
  \emph{Humanoid Robots, 2008. Humanoids 2008. 8th IEEE-RAS International
  Conference on}.\hskip 1em plus 0.5em minus 0.4em\relax IEEE, 2008, pp.
  406--412.

\bibitem{vs-2}
H.~Fujimoto, L.-C. Zhu, and K.~Abdel-Malek, ``Image-based visual servoing for
  grasping unknown objects,'' in \emph{Industrial Electronics Society, 2000.
  IECON 2000. 26th Annual Confjerence of the IEEE}, vol.~2.\hskip 1em plus
  0.5em minus 0.4em\relax IEEE, 2000, pp. 876--881.

\bibitem{levine2016end}
S.~Levine, C.~Finn, T.~Darrell, and P.~Abbeel, ``End-to-end training of deep
  visuomotor policies,'' \emph{The Journal of Machine Learning Research},
  vol.~17, no.~1, pp. 1334--1373, 2016.

\bibitem{multiview-pose}
A.~Zeng, K.-T. Yu, S.~Song, D.~Suo, E.~Walker, A.~Rodriguez, and J.~Xiao,
  ``Multi-view self-supervised deep learning for 6d pose estimation in the
  amazon picking challenge,'' in \emph{Robotics and Automation (ICRA), 2017
  IEEE International Conference on}.\hskip 1em plus 0.5em minus 0.4em\relax
  IEEE, 2017, pp. 1386--1383.

\bibitem{multiview-1}
C.~Mitash, K.~E. Bekris, and A.~Boularias, ``A self-supervised learning system
  for object detection using physics simulation and multi-view pose
  estimation,'' in \emph{Intelligent Robots and Systems (IROS), 2017 IEEE/RSJ
  International Conference on}.\hskip 1em plus 0.5em minus 0.4em\relax IEEE,
  2017, pp. 545--551.

\bibitem{yan2017sim}
M.~Yan, I.~Frosio, S.~Tyree, and J.~Kautz, ``Sim-to-real transfer of accurate
  grasping with eye-in-hand observations and continuous control,'' \emph{arXiv
  preprint arXiv:1712.03303}, 2017.

\bibitem{calli2018viewpoint}
B.~Calli, W.~Caarls, M.~Wisse, and P.~Jonker, ``Viewpoint optimization for
  aiding grasp synthesis algorithms using reinforcement learning,''
  \emph{Advanced Robotics}, pp. 1--13, 2018.

\bibitem{model-based}
S.~Gu, T.~Lillicrap, I.~Sutskever, and S.~Levine, ``Continuous deep q-learning
  with model-based acceleration,'' in \emph{International Conference on Machine
  Learning}, 2016, pp. 2829--2838.

\bibitem{model-free}
V.~Mnih, K.~Kavukcuoglu, D.~Silver, A.~Graves, I.~Antonoglou, D.~Wierstra, and
  M.~Riedmiller, ``Playing atari with deep reinforcement learning,''
  \emph{arXiv preprint arXiv:1312.5602}, 2013.

\bibitem{gu2016q}
S.~Gu, T.~Lillicrap, Z.~Ghahramani, R.~E. Turner, and S.~Levine, ``Q-prop:
  Sample-efficient policy gradient with an off-policy critic,'' \emph{arXiv
  preprint arXiv:1611.02247}, 2016.

\bibitem{gamrian2018transfer}
S.~Gamrian and Y.~Goldberg, ``Transfer learning for related reinforcement
  learning tasks via image-to-image translation,'' \emph{arXiv preprint
  arXiv:1806.07377}, 2018.

\bibitem{savva2019habitat}
M.~Savva, A.~Kadian, O.~Maksymets, Y.~Zhao, E.~Wijmans, B.~Jain, J.~Straub,
  J.~Liu, V.~Koltun, J.~Malik \emph{et~al.}, ``Habitat: A platform for embodied
  ai research,'' \emph{arXiv preprint arXiv:1904.01201}, 2019.

\bibitem{segd-1}
A.~Janoch, S.~Karayev, Y.~Jia, J.~T. Barron, M.~Fritz, K.~Saenko, and
  T.~Darrell, ``A category-level 3d object dataset: Putting the kinect to
  work,'' in \emph{Consumer depth cameras for computer vision}.\hskip 1em plus
  0.5em minus 0.4em\relax Springer, 2013, pp. 141--165.

\bibitem{segd-2}
N.~Silberman, D.~Hoiem, P.~Kohli, and R.~Fergus, ``Indoor segmentation and
  support inference from rgbd images,'' in \emph{European Conference on
  Computer Vision}.\hskip 1em plus 0.5em minus 0.4em\relax Springer, 2012, pp.
  746--760.

\bibitem{segd-3}
K.~Lai, L.~Bo, X.~Ren, and D.~Fox, ``A large-scale hierarchical multi-view
  rgb-d object dataset,'' in \emph{Robotics and Automation (ICRA), 2011 IEEE
  International Conference on}.\hskip 1em plus 0.5em minus 0.4em\relax IEEE,
  2011, pp. 1817--1824.

\bibitem{hand-data-1}
J.~Mccormac, A.~Handa, S.~Leutenegger, and A.~J. Davison, ``Scenenet rgb-d: 5m
  photorealistic images of synthetic indoor trajectories with ground truth.''
  \emph{arXiv: Computer Vision and Pattern Recognition}, 2016.

\bibitem{hand-data-2}
T.~Hodan, P.~Haluza, S.~Obdrzalek, J.~Matas, M.~I.~A. Lourakis, and X.~Zabulis,
  ``T-less: An rgb-d dataset for 6d pose estimation of texture-less objects,''
  \emph{workshop on applications of computer vision}, pp. 880--888, 2017.

\bibitem{small-data-1}
F.~Tombari, L.~D. Stefano, and S.~Giardino, ``Online learning for automatic
  segmentation of 3d data,'' pp. 4857--4864, 2011.

\bibitem{small-data-2}
A.~Richtsfeld, T.~Morwald, J.~Prankl, M.~Zillich, and M.~Vincze, ``Segmentation
  of unknown objects in indoor environments,'' pp. 4791--4796, 2012.

\bibitem{occd-1}
M.~Sun, G.~Bradski, B.-X. Xu, and S.~Savarese, ``Depth-encoded hough voting for
  joint object detection and shape recovery,'' in \emph{European Conference on
  Computer Vision}.\hskip 1em plus 0.5em minus 0.4em\relax Springer, 2010, pp.
  658--671.

\bibitem{occd-2}
S.~Hinterstoisser, V.~Lepetit, S.~Ilic, S.~Holzer, G.~Bradski, K.~Konolige, and
  N.~Navab, ``Model based training, detection and pose estimation of
  texture-less 3d objects in heavily cluttered scenes,'' in \emph{Asian
  conference on computer vision}.\hskip 1em plus 0.5em minus 0.4em\relax
  Springer, 2012, pp. 548--562.

\bibitem{redmon2016you}
J.~Redmon, S.~Divvala, R.~Girshick, and A.~Farhadi, ``You only look once:
  Unified, real-time object detection,'' in \emph{Proceedings of the IEEE
  conference on computer vision and pattern recognition}, 2016, pp. 779--788.

\bibitem{girshick2015fast}
R.~Girshick, ``Fast r-cnn,'' in \emph{Proceedings of the IEEE international
  conference on computer vision}, 2015, pp. 1440--1448.

\bibitem{ren2015faster}
S.~Ren, K.~He, R.~Girshick, and J.~Sun, ``Faster r-cnn: Towards real-time
  object detection with region proposal networks,'' in \emph{Advances in neural
  information processing systems}, 2015, pp. 91--99.

\bibitem{wang2019pseudo}
Y.~Wang, W.-L. Chao, D.~Garg, B.~Hariharan, M.~Campbell, and K.~Q. Weinberger,
  ``Pseudo-lidar from visual depth estimation: Bridging the gap in 3d object
  detection for autonomous driving,'' in \emph{Proceedings of the IEEE
  Conference on Computer Vision and Pattern Recognition}, 2019, pp. 8445--8453.

\bibitem{li2018pointcnn}
Y.~Li, R.~Bu, M.~Sun, W.~Wu, X.~Di, and B.~Chen, ``Pointcnn: Convolution on
  x-transformed points,'' in \emph{Advances in neural information processing
  systems}, 2018, pp. 820--830.

\bibitem{yang2018pixor}
B.~Yang, W.~Luo, and R.~Urtasun, ``Pixor: Real-time 3d object detection from
  point clouds,'' in \emph{Proceedings of the IEEE conference on Computer
  Vision and Pattern Recognition}, 2018, pp. 7652--7660.

\bibitem{zhu2017semantic}
Y.~Zhu, Y.~Tian, D.~Metaxas, and P.~Doll{\'a}r, ``Semantic amodal
  segmentation,'' in \emph{Proceedings of the IEEE Conference on Computer
  Vision and Pattern Recognition}, 2017, pp. 1464--1472.

\bibitem{li2016amodal}
K.~Li and J.~Malik, ``Amodal instance segmentation,'' in \emph{European
  Conference on Computer Vision}.\hskip 1em plus 0.5em minus 0.4em\relax
  Springer, 2016, pp. 677--693.

\bibitem{rl-app1}
V.~Mnih, A.~P. Badia, M.~Mirza, A.~Graves, T.~Lillicrap, T.~Harley, D.~Silver,
  and K.~Kavukcuoglu, ``Asynchronous methods for deep reinforcement learning,''
  in \emph{International conference on machine learning}, 2016, pp. 1928--1937.

\bibitem{rl-app2}
J.~Schulman, S.~Levine, P.~Abbeel, M.~Jordan, and P.~Moritz, ``Trust region
  policy optimization,'' in \emph{International Conference on Machine
  Learning}, 2015, pp. 1889--1897.

\bibitem{rl-app3}
S.~Levine and P.~Abbeel, ``Learning neural network policies with guided policy
  search under unknown dynamics,'' in \emph{Advances in Neural Information
  Processing Systems}, 2014, pp. 1071--1079.

\bibitem{gu2017deep}
S.~Gu, E.~Holly, T.~Lillicrap, and S.~Levine, ``Deep reinforcement learning for
  robotic manipulation with asynchronous off-policy updates,'' in \emph{2017
  IEEE international conference on robotics and automation (ICRA)}.\hskip 1em
  plus 0.5em minus 0.4em\relax IEEE, 2017, pp. 3389--3396.

\bibitem{wang2019tendencyrl}
C.~Wang, J.~Ding, X.~Chen, Z.~Ye, J.~Wang, Z.~Cai, and C.~Lu, ``Tendencyrl:
  Multi-stage discriminative hints for efficient goal-oriented reverse
  curriculum learning,'' in \emph{2019 IEEE/RSJ International Conference on
  Intelligent Robots and Systems (IROS)}.\hskip 1em plus 0.5em minus
  0.4em\relax IEEE, 2019, pp. 3474--3480.

\bibitem{sample-efficiency}
S.~Gu, T.~Lillicrap, Z.~Ghahramani, R.~E. Turner, and S.~Levine, ``Q-prop:
  Sample-efficient policy gradient with an off-policy critic,'' \emph{arXiv
  preprint arXiv:1611.02247}, 2016.

\bibitem{levine2018learning}
S.~Levine, P.~Pastor, A.~Krizhevsky, J.~Ibarz, and D.~Quillen, ``Learning
  hand-eye coordination for robotic grasping with deep learning and large-scale
  data collection,'' \emph{The International Journal of Robotics Research},
  vol.~37, no. 4-5, pp. 421--436, 2018.

\bibitem{hrl}
T.~G. Dietterich, ``Hierarchical reinforcement learning with the maxq value
  function decomposition,'' \emph{Journal of artificial intelligence research},
  vol.~13, pp. 227--303, 2000.

\bibitem{il-ap1}
J.~Ho and S.~Ermon, ``Generative adversarial imitation learning,'' in
  \emph{Advances in neural information processing systems}, 2016, pp.
  4565--4573.

\bibitem{il-ap2}
S.~Reddy, A.~D. Dragan, and S.~Levine, ``Sqil: Imitation learning via
  regularized behavioral cloning,'' \emph{arXiv preprint arXiv:1905.11108},
  2019.

\bibitem{il-app1}
A.~Nair, D.~Chen, P.~Agrawal, P.~Isola, P.~Abbeel, J.~Malik, and S.~Levine,
  ``Combining self-supervised learning and imitation for vision-based rope
  manipulation,'' in \emph{2017 IEEE International Conference on Robotics and
  Automation (ICRA)}.\hskip 1em plus 0.5em minus 0.4em\relax IEEE, 2017, pp.
  2146--2153.

\bibitem{il-app2}
Y.~Liu, A.~Gupta, P.~Abbeel, and S.~Levine, ``Imitation from observation:
  Learning to imitate behaviors from raw video via context translation,'' in
  \emph{2018 IEEE International Conference on Robotics and Automation
  (ICRA)}.\hskip 1em plus 0.5em minus 0.4em\relax IEEE, 2018, pp. 1118--1125.

\bibitem{il}
A.~Attia and S.~Dayan, ``Global overview of imitation learning,'' \emph{arXiv
  preprint arXiv:1801.06503}, 2018.

\bibitem{il-1}
C.~Finn, T.~Yu, T.~Zhang, P.~Abbeel, and S.~Levine, ``One-shot visual imitation
  learning via meta-learning,'' \emph{arXiv preprint arXiv:1709.04905}, 2017.

\bibitem{il-2}
Y.~Duan, M.~Andrychowicz, B.~Stadie, O.~J. Ho, J.~Schneider, I.~Sutskever,
  P.~Abbeel, and W.~Zaremba, ``One-shot imitation learning,'' in \emph{Advances
  in neural information processing systems}, 2017, pp. 1087--1098.

\bibitem{il-hil}
T.~Yu, P.~Abbeel, S.~Levine, and C.~Finn, ``One-shot hierarchical imitation
  learning of compound visuomotor tasks,'' \emph{arXiv preprint
  arXiv:1810.11043}, 2018.

\bibitem{motion-planning-1}
J.~Kuffner, K.~Nishiwaki, S.~Kagami, M.~Inaba, and H.~Inoue, ``Motion planning
  for humanoid robots,'' in \emph{Robotics Research. The Eleventh International
  Symposium}.\hskip 1em plus 0.5em minus 0.4em\relax Springer, 2005, pp.
  365--374.

\bibitem{petrelli2011repeatability}
A.~Petrelli and L.~Di~Stefano, ``On the repeatability of the local reference
  frame for partial shape matching,'' in \emph{Computer Vision (ICCV), 2011
  IEEE International Conference on}.\hskip 1em plus 0.5em minus 0.4em\relax
  IEEE, 2011, pp. 2244--2251.

\bibitem{papazov2010efficient}
C.~Papazov and D.~Burschka, ``An efficient ransac for 3d object recognition in
  noisy and occluded scenes,'' in \emph{Asian Conference on Computer
  Vision}.\hskip 1em plus 0.5em minus 0.4em\relax Springer, 2010, pp. 135--148.

\bibitem{realsense}
``Intel realsense d435,'' \url{
  https://click.intel.com/intelr-realsensetm-depth-camerad435.html}.

\bibitem{gpd-1}
A.~ten Pas, M.~Gualtieri, K.~Saenko, and R.~Platt, ``Grasp pose detection in
  point clouds,'' \emph{The International Journal of Robotics Research},
  vol.~36, no. 13-14, pp. 1455--1473, 2017.

\bibitem{lenet}
Y.~LeCun, L.~Bottou, Y.~Bengio, and P.~Haffner, ``Gradient-based learning
  applied to document recognition,'' \emph{Proceedings of the IEEE}, vol.~86,
  no.~11, pp. 2278--2324, 1998.

\bibitem{bigbird}
A.~Singh, J.~Sha, K.~S. Narayan, T.~Achim, and P.~Abbeel, ``Bigbird: A
  large-scale 3d database of object instances,'' in \emph{Robotics and
  Automation (ICRA), 2014 IEEE International Conference on}.\hskip 1em plus
  0.5em minus 0.4em\relax IEEE, 2014, pp. 509--516.

\bibitem{ma2017multi}
L.~Ma, J.~St{\"u}ckler, C.~Kerl, and D.~Cremers, ``Multi-view deep learning for
  consistent semantic mapping with rgb-d cameras,'' in \emph{2017 IEEE/RSJ
  International Conference on Intelligent Robots and Systems (IROS)}.\hskip 1em
  plus 0.5em minus 0.4em\relax IEEE, 2017, pp. 598--605.

\end{thebibliography}

\end{document}